\documentclass[10pt,twocolumn,letterpaper]{article}

\usepackage{cvpr}
\usepackage{times}
\usepackage{epsfig}
\usepackage{graphicx}
\usepackage{amsmath}
\usepackage{amssymb}

\usepackage[pagebackref=true,breaklinks=true,letterpaper=true,colorlinks,bookmarks=false]{hyperref}

\cvprfinalcopy 

\def\cvprPaperID{12} 

\ifcvprfinal\fi
\begin{document}

\title{Leveraging Context to Support Automated Food Recognition in Restaurants}

\author{
  Vinay Bettadapura, Edison Thomaz, Aman Parnami, Gregory D. Abowd, Irfan Essa
  \\
  \\
  School of Interactive Computing
  \\
  Georgia Institute of Technology, Atlanta, Georgia, USA
  \and
    \href{http://www.vbettadapura.com/egocentric/food}{\small http://www.vbettadapura.com/egocentric/food}
}

\maketitle
\thispagestyle{empty}

\begin{abstract}

The pervasiveness of mobile cameras has resulted in a dramatic increase in food photos, which are pictures reflecting what people eat.  In this paper, we study how taking pictures of what we eat in restaurants can be used for the purpose of automating food journaling. We propose to leverage the context of where the picture was taken, with additional information about the restaurant, available online, coupled with state-of-the-art computer vision techniques to recognize the food being consumed. To this end, we demonstrate image-based recognition of foods eaten in restaurants by training a classifier with images from restaurant's online menu databases. We evaluate the performance of our system in unconstrained, real-world settings with food images taken in 10 restaurants across 5 different types of food (American, Indian, Italian, Mexican and Thai).
 
\end{abstract}

\section{Introduction}

Recent studies show strong evidence that adherence to dietary self-monitoring helps people lose weight and meet dietary goals \cite{Burke:2011fn}. This is critically important since obesity is now a major public health concern associated with rising rates of chronic disease and early death \cite{doi:10.1001/jama.293.15.1861}. 

Although numerous methods have been suggested for addressing the problem of poor adherence to nutrition journaling \cite{Amft:2008bs,Sazonov:2008fi,Yatani:2012ux}, a truly practical system for objective dietary monitoring has not yet been realized; the most common technique for logging eating habits today remains self-reports through paper diaries and more recently, smartphone applications. This process is tedious, time-consuming, prone to errors and leads to selective under reporting \cite{gorisUnderReporting}. 

While needs for automated food journaling persist, we are seeing an ever increasing growth in people photographing what they eat. In this paper we present a system and approach for automatically recognizing foods eaten at restaurants from first-person food photos with the goal of facilitating food journaling. The methodology we employ is unique because it leverages sensor data (i.e., location) captured at the time photos are taken. Additionally, online resources such as restaurant menus and online images are used to help recognize foods once a location has been identified. 

Our motivation for focusing on restaurant eating activities stems from findings from recent surveys indicating a trend towards eating out versus eating at home. In 1970, 25.9 percent of all food spending was on food away from home; by 2012, that share rose to its highest level of 43.1 percent \cite{USDA}. Additionally, 8 in 10 Americans report eating at fast-food restaurants at least monthly, with almost half saying they eat fast food at least weekly \cite{GALLUP}.

Research in the computer vision community has explored the recognition of either a small sub-set of food types in controlled laboratory environments \cite{PFID,pairWiseFood} or food images obtained from the web \cite{japanFood}. However, there have been only a few validated implementations that address the challenge of food recognition from images taken ``in the wild'' \cite{Kitamura:2010dl}. Systems that rely on crowdsourcing, such as PlateMate \cite{Noronha:2011vs}, have shown promise but are limited in terms of cost and scalability. Additionally, privacy concerns might arise when food photographs are reviewed by untrusted human computation workers \cite{Thomaz:2013iv}. 

In this paper, we seek an approach that supports automatic recognition of food, leveraging the context of where the photograph was taken. Our contributions are:

\begin{figure*}
\begin{center}
\includegraphics[height=5.5cm]{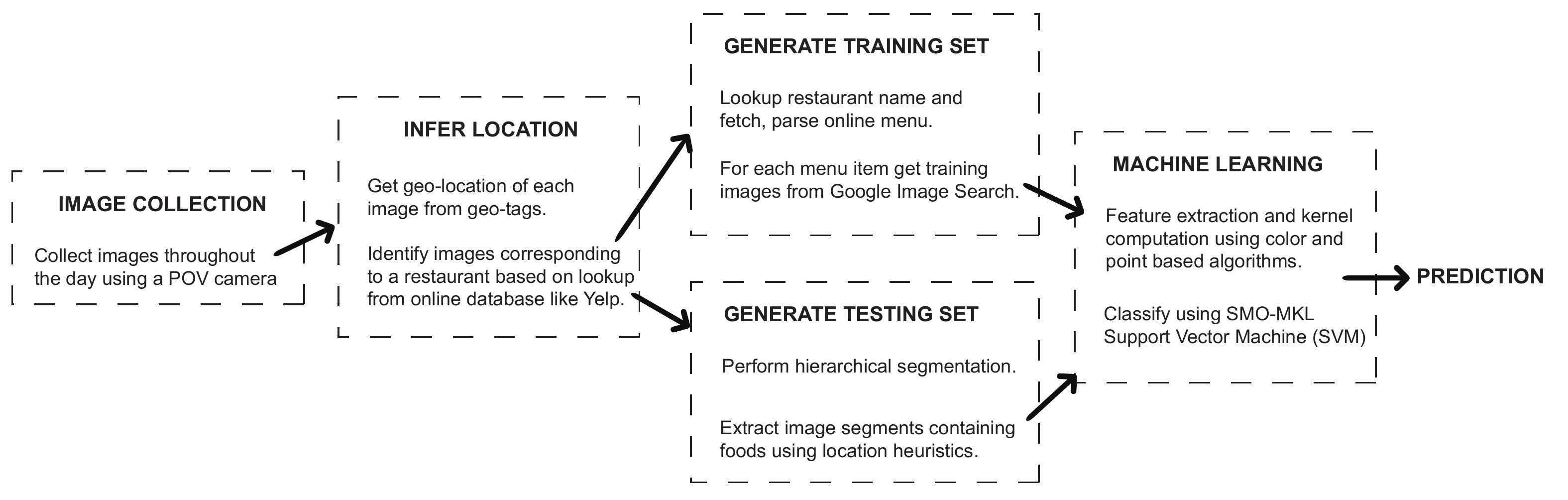}
\end{center}
\caption{An overview of our automatic food recognition approach.}
\label{fig:Block-diagram}
\end{figure*}

\begin{itemize}

\item An automatic workflow where online resources are queried with contextual sensor data to find food images and additional information about the restaurant where the food picture was taken, with the intent to build classifiers for food recognition.

\item An image classification approach using the SMO-MKL multi-class SVM classification framework with features extracted from test photographs. 

\item An in-the-wild evaluation of our approach with food images taken in 10 restaurants across 5 different types of cuisines (American, Indian, Italian, Mexican and Thai). 

\item A comparative evaluation focused on the effect of location data in food recognition results.

\end{itemize}

In this paper, we concentrate on food recognition, leveraging the additional context that is available (location, websites, etc.). Our goal in this paper is to in essence, using food and restaurants as the domain, demonstrate the value of external context, coupled with image recognition to support classification. We believe that the same method can be used for many other domains.

\section{Related Work}

Various sensor-based methods for automated dietary monitoring have been proposed over the years. Amft and Troster~\cite{Amft:2008bs} explored sensors in the wrists, head and neck to automatically detect food intake gestures, chewing, and swallowing from accelerometer and acoustic sensor data. Sazonov et al.~built a system for monitoring swallowing and chewing using a piezoelectric strain gauge positioned below the ear and a small microphone located over the laryngopharynx \cite{Sazonov:2008fi}. Yatani and Truong presented a wearable acoustic sensor attached to the user's neck \cite{Yatani:2012ux} while Cheng et al.~explored the use of a neckband for nutrition monitoring \cite{Cheng:2013ex}.

With the emergence of low-cost, high-resolution wearable cameras, recording individuals as they perform everyday activities such as eating has been gaining appeal \cite{Arab:2011gj}. In this approach, individuals wear cameras that take first-person point-of-view photographs periodically throughout the day. Although first-person point-of-view images offer a viable alternative to direct observation, one of the fundamental problems is image analysis. All captured images must be manually coded for salient content (e.g., evidence of eating activity), a process tends to be tedious and time-consuming. 

Over the past decade, research in computer vision is moving towards ``in the wild'' approaches. Recent research has focussed on recognizing realistic actions in videos \cite{ActionsInWild}, unconstrained face verification and labeling \cite{NeerajAttribute} and objection detection and recognition in natural images \cite{PASCAL-VOC}. Food recognition in the wild using vision-based methods is growing as a topic of interest, with Kitamura et al.~\cite{Kitamura:2010dl} showing promise. 

Finally, human computation lies in-between completely manual and fully-automated vision-based image analysis. PlateMate \cite{Noronha:2011vs} crowdsources nutritional analysis from food photographs using Amazon Mechanical Turk, and Thomaz et al.~investigated the use of crowdsourcing to detect \cite{Thomaz:2013kc} eating moments from first-person point-of-view images. Despite the promise of these crowdsourcing-based approaches, there are clear benefits to a fully automated method in economic terms, and possibly with regards to privacy as well.

\begin{figure*}
\begin{centering}
\includegraphics[height=4.5cm]{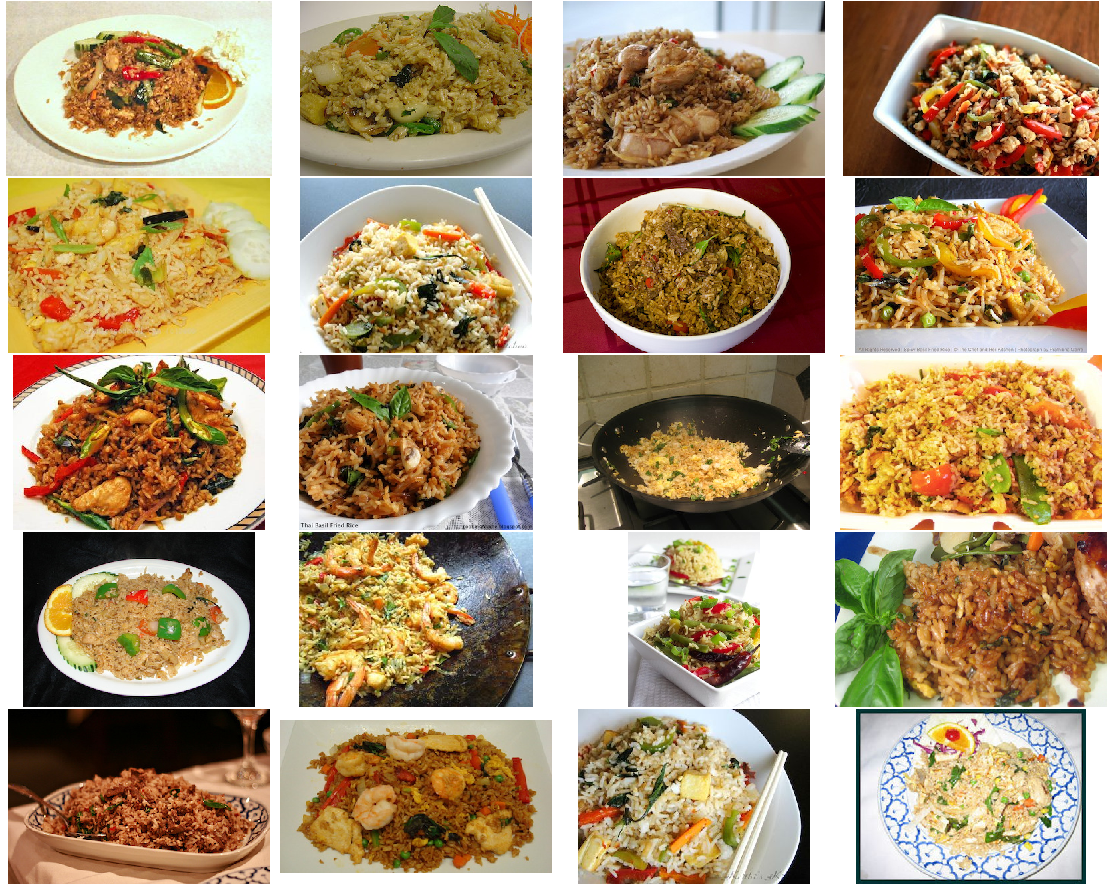}
\hspace{1em}
\includegraphics[height=4.5cm]{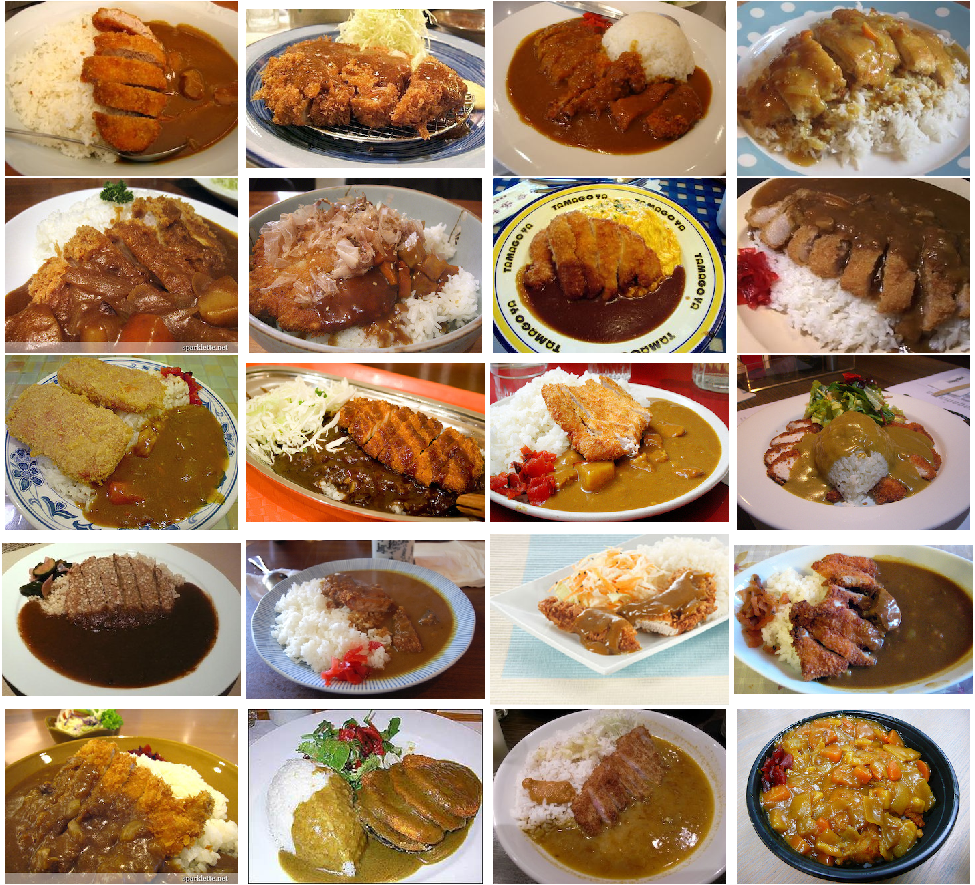}
\hspace{1em}
\includegraphics[height=4.5cm]{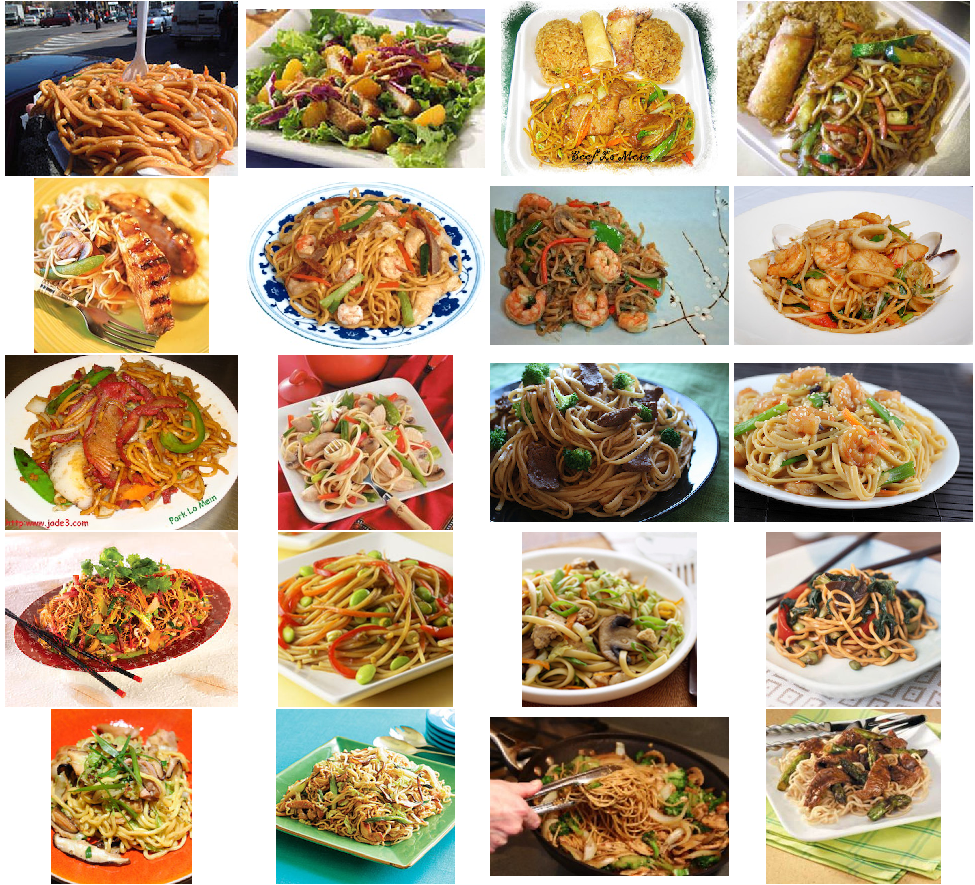}
\par\end{centering}
\caption{\label{fig:Training-images}Weakly-labeled training images obtained from Google Image search for 3 classes of food: \textbf{Left:} Basil Fried Rice; \textbf{Center:} Curry Katsu; \textbf{Right:} Lo Mein.}
\end{figure*}

\section{Methodology}

Recognizing foods from photographs is a challenging undertaking. The complexity arises from the large number of food categories, variations in their appearance and shape, the different ways in which they are served and the environmental conditions they are presented in. To offset the difficulty of this task, the methodology we propose (Figure \ref{fig:Block-diagram}) centers on the use of location information about the eating activity, and also restaurant menu databases that can be queried online. As noted, our technique is specifically aimed at eating activities in restaurants as we leverage the context of restaurant related information for classification.

\subsection{Image Acquisition}

The first step in our approach involves the acquisition of food images. The popularity of cameras in smartphones and wearable devices like Google Glass makes it easy to capture food images in restaurants. In fact, many food photographs communities such as FoodGawker have emerged over the last several years, all centered on food photo sharing. Photographing food is also hitting major photo sharing sites like Instagram, Pinterest and Flickr, and food review sites like Yelp. These food-oriented photo activities illustrate the practicality of using manually-shot food photos for food recognition.

\subsection{Geo-Localizing Images}

The second step involves associating food photos with longitude and latitude coordinates. If the camera that is being used supports image geo-tagging, then the process of localizing images is greatly simplified. Commodity smart-phones and cameras like the Contour and SenseCam come with built-in GPS capabilities. If the geo-tag is not available, image localization techniques can be used \cite{ZamirStreetView}. Once location is obtained for all captured images, the APIs of Yelp and Google Places are valuable for matching the images' geo-tags coincide with the geo-tag of a restaurant.

\subsection{Weakly Supervised Learning}

Being able to localize images to a restaurant greatly constrains the problem of food classification in the wild. A strong assumption can be made that the food present in the images must be from one of the items on the restaurant's menu. This key observation makes it possible to build a weakly supervised classification framework for food classification. The subsequent sections describe in detail the gathering of weakly-labeled training data, preparing the test data and classification using the SMO-MKL multi-class SVM classification framework \cite{SMO-MKL}.

\subsubsection{Gathering Training Data}

We start with collecting images localized to a particular restaurant $R$. Once we know $R$, we can use the web as a knowledge-base and search for $R$'s menu. This task is greatly simplified thanks to online data-sources like Yelp, Google Places, Allmenus.com and Openmenu.com, which provides comprehensive databases of restaurant menus.

Let the menu for $R$ be denoted by $M_{R}$ and let the items on the menu be $m_{i}$. For each $m_{i}\in M_{R}$, the top 50 images of $m_{i}$ are downloaded using search engines like Google Image search. This comprises the weakly-labeled training data. Three examples are shown in Figure \ref{fig:Training-images}. From the images, it is possible to see that there is a high degree of intra-class variability in terms of color and presentation of food. As is the case with any state-of-the-art object recognition system, our approach relies on the fact that given sufficient number of images for each class, it should be possible to learn common patterns and statistical similarities from the images.

\subsubsection{Preparing Testing Data}

\begin{figure}
\begin{centering}
\includegraphics[width=0.85\columnwidth]{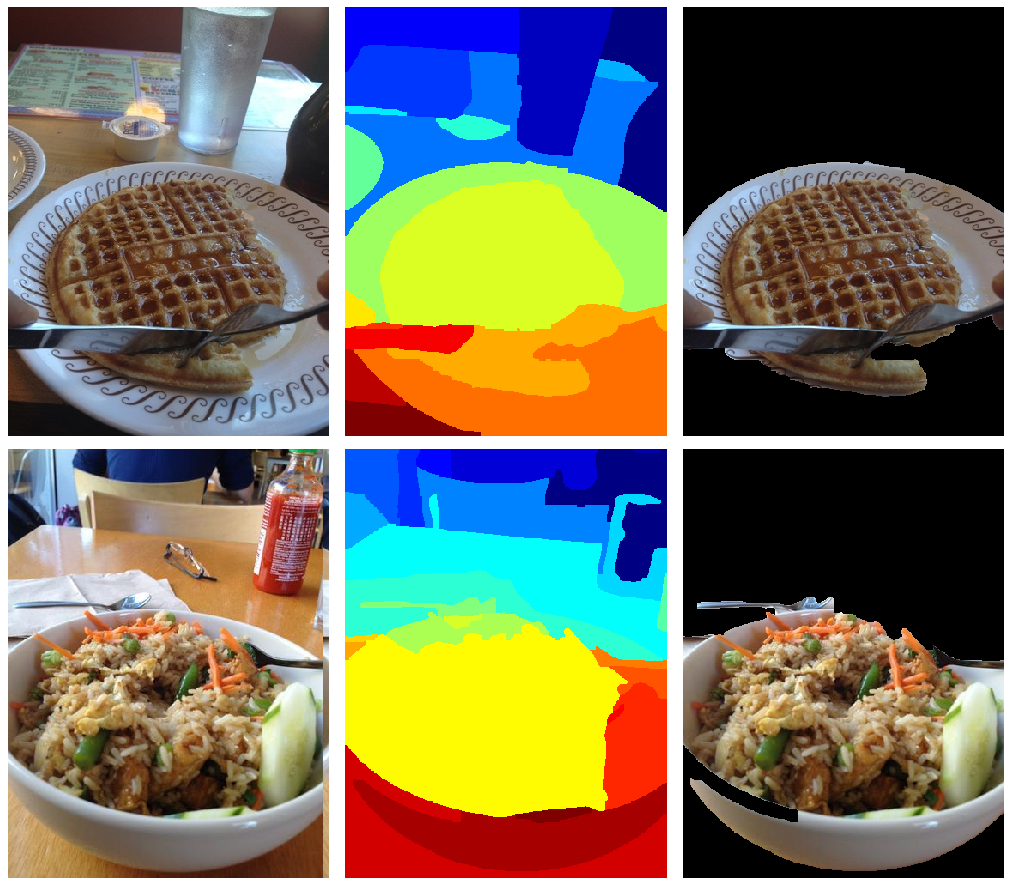}
\par\end{centering}
\caption{\label{fig:Extracting-image-segments}Extracting segments using hierarchical segmentation. The final segmented image is shown on the right.}
\end{figure}

The test images, localized to restaurant $R$, are segmented using hierarchical segmentation and the segments are extracted from parts of the image where we expect the food to be present \cite{segmentation}. The final set of segmented images forms our test data. An example is shown in Figure \ref{fig:Extracting-image-segments}.

\subsubsection{Feature Descriptors}

Choosing the right combination of feature detectors, descriptors and classification backend is key to achieving good accuracy in any object recognition or image categorization task. While salient point detectors and corresponding region descriptors can robustly detect regions which are invariant to translation, rotation and scale \cite{SIFT,AffineDetectorsComparison}, illumination changes can still cause performance to drop. This is a cause of concern when dealing with food images, since images taken at restaurants are typically indoors and under varying lighting conditions. Recent work by van de Sande et al.~\cite{colorDescriptors} systematically studies the invariance properties and distinctiveness of color descriptors. The results of this study guided the choice of the descriptors in our approach. For the classification back-end, we use Multiple Kernel Learning (MKL), which in recent years, has given robust performance on object categorization
tasks~\cite{MKL-Jordan,MKL-sonnenburg,SMO-MKL}.

For feature extraction from the training and test data, a Harris-Laplace point detector is used since it has shown good performance for category recognition tasks \cite{chi-Square} and is scale-invariant. However the choice of feature descriptor is more complicated. As seen in Figure \ref{fig:Training-images}, there is a high degree of intra-class variability in terms of color and lighting. Based on the recent work by van de Sande et al.~\cite{colorDescriptors} that studies the invariance properties and distinctiveness of various color descriptors on light intensity and color changes, we pick the following six descriptors, 2 color-based and 4 SIFT-based (Scale-Invariant Feature Transform~\cite{SIFT}):

\textbf{Color Moment Invariants: }Generalized color moments $M_{pq}^{abc}$ (of order $p+q$ and degree $a+b+c$) have been defined as $M_{pq}^{abc}=\int\int x^{p}y^{q}[I_{R}(x,y)]^{a}[I_{G}(x,y)]^{b}[I_{B}(x,y)]^{c}dxdy$. Color moment invariants are those combinations of generalized color moments that allow for normalization against photometric changes and are invariant to changes and shifts in light intensity and color.

\textbf{Hue Histograms: }Based on the observation that the certainty of hue is inversely proportional to the saturation, each hue sample in the hue histogram is weighted by its saturation. This helps overcome the (known) instability of hue near the gray axis in HSV space. The descriptors obtained are invariant to changes and shifts in light intensity.

\textbf{C-SIFT:} The descriptors are built using the C-invariant (normalized opponent color space). C-SIFT is invariant to changes in light intensity.

\textbf{OpponentSIFT: }All the channels in the opponent color space are described using SIFT descriptors. They are invariant to changes and shifts in light intensity.

\textbf{RGB-SIFT: }SIFT descriptors are computed for every RGB channel independently. The resulting descriptors are invariant to changes and shifts in light intensity and color.

\textbf{SIFT: }The original SIFT descriptor proposed by Lowe \cite{SIFT}. It is invariant to changes and shifts in light intensity.

\subsubsection{Classification Using SMO-MKL}

For a given restaurant $R$, 100,000 interest points are detected in the training data and for each of the 6 descriptors, visual codebooks are built using $k$-means clustering with $k$ = 1000. Using these codebooks, bag-of-words (BoW) histograms are built for the training images. Similarly, interest points are detected in the test images and BoW are built for the 6 descriptors (using the visual codebooks generated with the training data).

For each of the 6 sets of BoW features, extended Gaussians kernels of the following form are computed:

\begin{equation}
K(H_{i},H_{j})=\exp(-\frac{1}{A}D(H_{i},H_{j}))
\end{equation}

where $H_{i}=\{h_{in}\}$ and $H_{j}=\{h_{jn}\}$ are the BoW histograms (scaled between 0 to 1 such that they lie within a unit hypersphere) and $D(H_{i},H_{j})$ is the $\chi^{2}$ distance defined as

\begin{equation}
D(H_{i},H_{j})=\frac{1}{2}\sum_{n=1}^{V}\frac{(h_{in}-h_{jn})^{2}}{h_{in}+h_{jn}}
\end{equation}

where $V$ is the vocabulary size (1000, in our case). The parameter $A$ is the mean value of the distances between all the training examples \cite{chi-Square}. Given the set of these $N$ base kernels $\{K_{k}\}$ (in our case $N$ = 6), linear MKL aims to learn a linear combination of the base kernels: $K=\sum_{k=1}^{N}\alpha_{i}K_{i}$

But the standard MKL formulation subject to $l_{1}$ regularization leads to a dual that is not differentiable. Hence the Sequential Minimal Optimization (SMO) algorithm cannot be applied and more expensive alternatives have to be pursued. Recently, Vishwanathan et al.\ showed that it is possible to use the SMO algorithm if the focus is on training $p$-norm MKL, with $p>1$ \cite{SMO-MKL}. They also show that the SMO-MKL algorithm is robust and significantly faster than the state-of-the-art $p$-norm MKL solvers. In our experiments, we train and test using the SMO-MKL SVM.

\section{Study \& Evaluation}

We perform two sets of experiments to evaluate our approach. In the first set of experiments, we compare the feature extraction and classification techniques used in this paper, with the state-of-the-art food recognition algorithms on the PFID benchmark data-set \cite{PFID}. This validates our proposed approach. In the second set of experiments, we measure the performance of the proposed approach for ``in-the-wild'' food recognition.

\subsection{Comparative Evaluations}
\begin{figure}
\begin{centering}
\includegraphics[width=1.0\columnwidth]{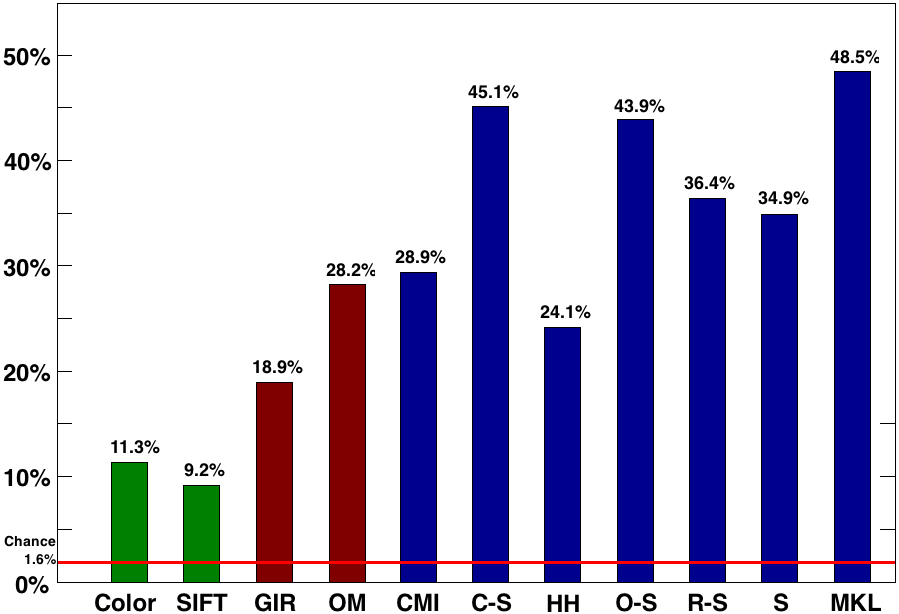}
\par\end{centering}
\caption{\label{fig:pfid-comparison}Performance of the 6 feature descriptors and SMO-MKL on the PFID data-set. The first two results (shown in green) are the baseline for PFID published by \cite{PFID}. The next two (shown in red) are the results obtained by using Global Ingredient Representation (GIR) and Orientation and Midpoint Category (OM) \cite{pairWiseFood}. The rest of the results (in blue) are one ones obtained using the 6 feature descriptors and MKL (CMI: Color Moment Invariant, C-S: C-SIFT, HH: Hue-Histogram, O-S: OpponentSIFT, R-S: RGB- SIFT, S: SIFT and MKL: Multiple Kernel Learning). MKL gives the best performance on this data-set.}
\end{figure}

We study the performance of the 6 feature descriptors and SMO-MKL classification on the PFID food data-set. The PFID dataset is a collection of 61 categories of fast food images acquired under lab conditions. Each category contains 3 different instances of food with 6 images from 6 view-points in each instance. In order to compare our results with the previous published results on PFID \cite{PFID,pairWiseFood}, we follow the same protocol used by them, i.e. a 3-fold cross-validation is performed with 12 images from one instance being used for training while the other 6 images from the remaining instance are used for testing.
The results of our experiments are shown in Figure \ref{fig:pfid-comparison}. MKL gives the best performance and improves the state-of- the-art \cite{pairWiseFood} by more than 20\%. It is interesting to note that the SIFT descriptor used in our approach achieves 34.9\% accuracy whereas the SIFT descriptor used in the PFID baseline \cite{PFID} achieves 9.2\% accuracy. The reason for this difference is that the authors of the PFID baseline use LIB-SVM for classification with it’s default parameters. However, by switching to the $\chi^2$ kernel (and ensuring that the data is scaled) and by tuning the SVM parameters (through a grid-search over the space of $C$ and $\gamma$), we can get a significant boost in performance with just SIFT features alone.

\subsection{Food Recognition in Restaurants}

\begin{figure}
\begin{centering}
\includegraphics[width=0.95\columnwidth]{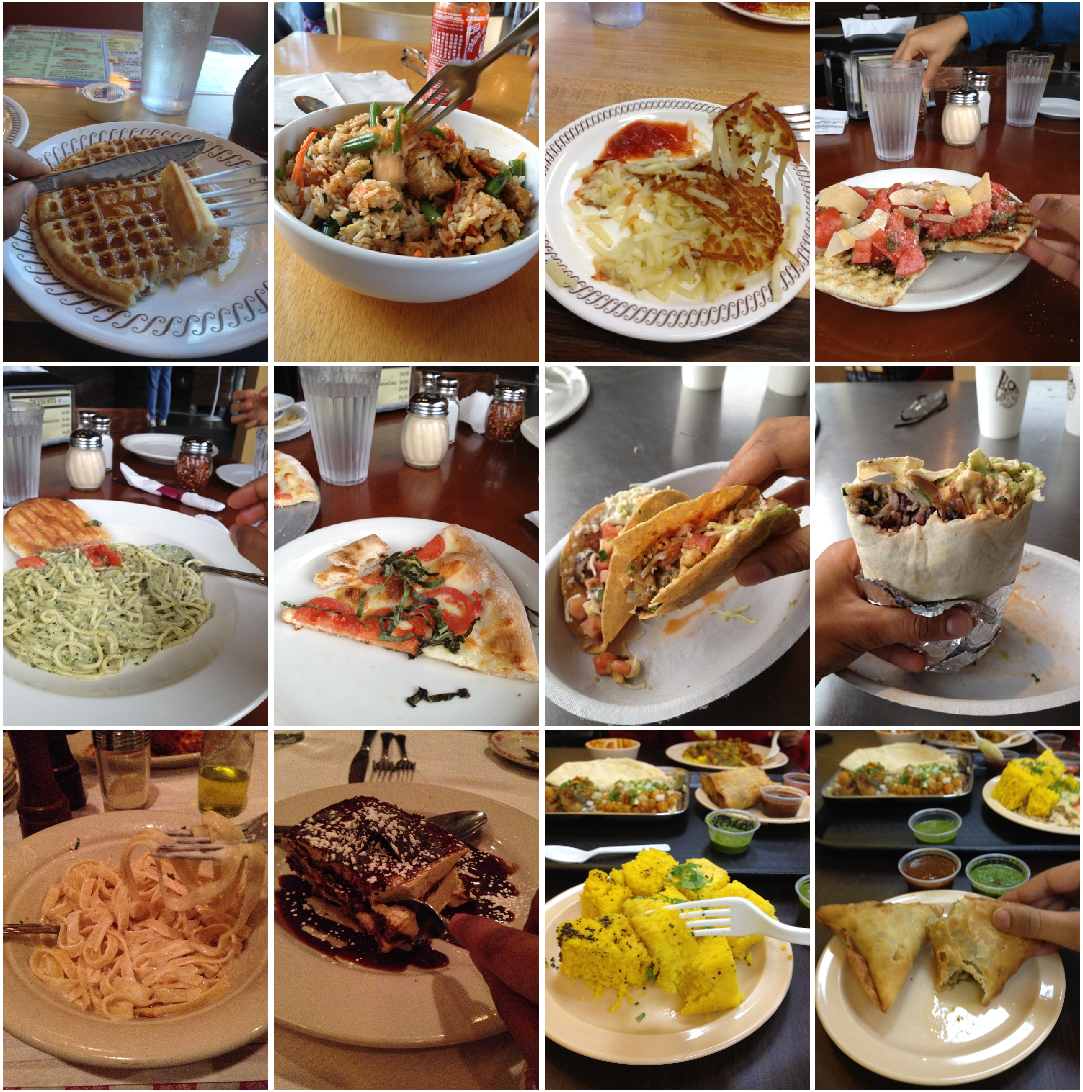}
\par\end{centering}
\caption{\label{fig:Sample-Test-Images}Sample (12 out of 600) of the ``in-the-wild'' images used in testing.}
\end{figure}

\begin{figure}
\begin{centering}
\includegraphics[width=0.95\columnwidth]{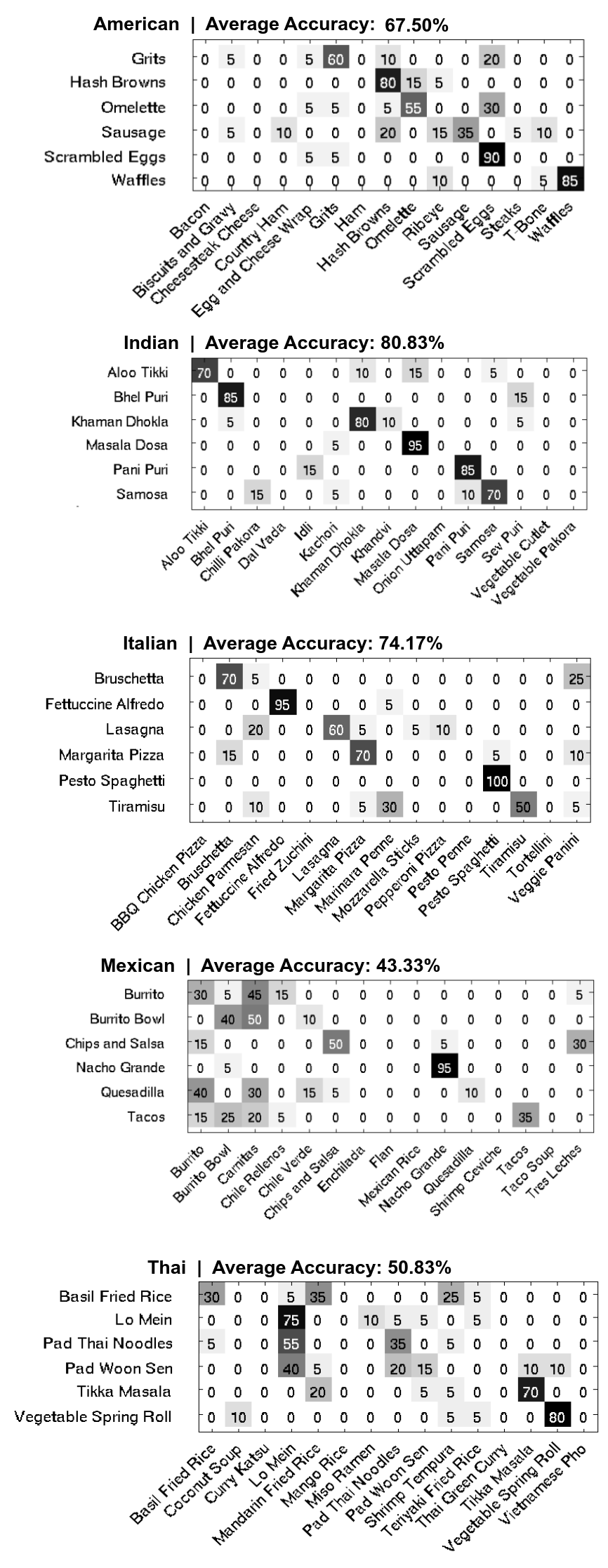}
\par\end{centering}
\caption{\label{fig:Confusion-matrices}Confusion matrices for the best performing features of Table \ref{tab:Results-table} (for each of the 5 cuisines). Darker colors show better recognition. The 6 food classes in the rows are the ones used for testing and the 15 food classes in the columns are the ones used for training. The overall average accuracy is 63.33\%}
\end{figure}

To study the performance and the practicality of our approach, experiments were conducted on images collected from restaurants across 5 different cuisines: American, Indian, Italian, Mexican and Thai. To discount for user and location bias, 3 different individuals collected images on different days from 10 different restaurants (2 per cuisines). The data collection was done in two phases. In the first phase, the food images were captured using smartphone cameras. In total, 300 ``in-the-wild'' food images (5 cuisines $\times$ 6 dishes/cuisine $\times$ 10 images/dish) were obtained. In the second phase, data collection was repeated using a Google Glass and an additional 300 images were captured. These 600 ``in-the-wild'' images, form our test data-set. A sample of these test images is shown in Figure \ref{fig:Sample-Test-Images}.

Using the geo-location information, the menu for each restaurant was automatically retrieved. For our experiments, we restricted the training to 15 dishes from each cuisine (selected based on online popularity). For each of the 15 dishes on the menu, 50 training images were downloaded using Google Image search. Thus, a total of 3,750 weakly-labeled training images were downloaded (5 cuisines $\times$ 15 menu-items/cuisine $\times$ 50 training-images/menu-item).

\begin{table*}
\begin{centering}
\scalebox{0.9}{
\begin{tabular}{|c|c|c|c|c|c|c|c|}
\hline 
 & {\small CMI} & {\small C-S} & {\small HH} & {\small O-S} & {\small R-S} & {\small S} & {\small MKL}\tabularnewline
\hline 
\hline 
{\small American} & {\small 45.8\%} & {\small 51.7\%} & {\small 43.3\%} & {\small 43.3\%} & {\small 37.5\%} & {\small 29.2\%} & \textbf{\small 67.5\%}\tabularnewline
\hline 
{\small Indian} & {\small 44.2\%} & {\small 74.2\%} & {\small 55.0\%} & {\small 59.2\%} & {\small 69.2\%} & {\small 65.0\%} & \textbf{\small 80.8\%}\tabularnewline
\hline 
{\small Italian} & {\small 33.3\%} & {\small 52.5\%} & {\small 67.5\%} & \textbf{\small 74.2\%} & {\small 66.7\%} & {\small 49.2\%} & {\small 67.5\%}\tabularnewline
\hline 
{\small Mexican} & {\small 36.7\%} & {\small 35.8\%} & {\small 20.8\%} & {\small 37.5\%} & {\small 24.2\%} & {\small 33.3\%} & \textbf{\small 43.3\%}\tabularnewline
\hline 
{\small Thai} & {\small 27.5\%} & {\small 36.7\%} & {\small 25.0\%} & {\small 33.3\%} & \textbf{\small 50.8\%} & {\small 30.8\%} & \textbf{\small 50.8\%}\tabularnewline
\hline 
\end{tabular}}
\par\end{centering}

\medskip{}

\caption{\label{tab:Results-table}Classification results showing the performance of the various feature descriptors on the 5 cuisines. The columns are: CMI: Color Moment Invariant, C-S: C-SIFT, HH: Hue-Histogram, O-S: OpponentSIFT, R-S: RGB-SIFT, S: SIFT and MKL: Multiple Kernel Learning.}

\end{table*}

Next, we perform interest point detection, feature extraction, codebook building for BoW representation, kernel pre-computation and finally classification using SMO-MKL. The results are summarized in Table \ref{tab:Results-table} and the individual confusion matrices are shown in Figure \ref{fig:Confusion-matrices}. We achieve good classification accuracy with American, Indian and Italian cuisines. However, for the Mexican and Thai cuisines, the accuracy is limited. It could be due to the fact that there is a low degree of visible variability between food types belonging to the same cuisines. For example, in the confusion matrix for Thai, we can see that Basil Fried Rice is confused with Mandarin Fried Rice and Pad Thai Noodles is confused with Lo Mein. It could be very hard, even for humans, to distinguish between such classes by looking at their images.

From Table \ref{tab:Results-table}, we can see that there is no single descriptor that works well across all the 5 cuisines. This could be due to the high-degree of variation in the training data. However, combining the descriptors using MKL yields the best performance in 4 out of the 5 cases.

\subsection{Recognition Without Location Prior}

Our approach is based on the hypothesis that knowing the location (through geo-tags) helps us in narrowing down the number of food categories which in turn boosts recognition rates. In order to test this hypothesis, we disregard the location information and train our SMO-MKL classifier on all of the training data (3,750 images). With this setup, accuracy across our 600 test images is 15.67\%. On the other hand, the overall average accuracy across the 5 cuisines (from Figure \ref{fig:Confusion-matrices}) is 63.33\%. We can see that the average performance increased by 47.66\% when location prior was included. This provides validation that knowing the location of eating activities helps in food recognition, and that it is better to build several smaller restaurant/cuisine specific classifiers rather than one all-category food classifier.

\section{Discussion}

In this section we discuss several important points pertaining to the generalizability of our approach, implementation issues, and practical considerations.

\paragraph{Generalizability} The automatic food identification approach that we propose is focused on eating activities in restaurants. Although this might seem limiting, eating out has been growing in popularity and 43.1\% of food spending was reported to having been spent in foods away from home in 2012 \cite{GALLUP,USDA}. Moreover, we feel that eating and food information gathered in restaurants is more valuable for dietary self-monitoring than food information obtained at home, since individuals are more likely to know food types and the composition of food items prepared in their own homes.

We designed our study and evaluation with the goal of maximizing the external validity of our results. We evaluated our approach by having three individuals collect images from the most popular restaurant types by cuisine in the US on different days and using two different devices (smartphones and Google Glass). We feel confident that our methodology will scale in the future, especially since it leverages sensor data, online resources and practices around food imagery that will become increasingly more prevalent in years to come.

One important aspect of the approach is that it depends on weakly-labeled training images obtained from the web. The high-degree of intra-class variability for the same food across different restaurants has a negative effect on performance. A promising alternative is to train on (automatically acquired) food images taken at the same restaurant as the one where the test images were taken. While getting this kind of data seems difficult, it may soon be possible. A recently launched service by Yelp (among others), allows users to upload photos of their food. With such crowd-sourced imagery available for a given restaurant, it may soon be possible to train specialized classifiers for that restaurant. In our future work, we plan to test this hypothesis and improve the recognition accuracies.

\paragraph{Location Error}

We not only identify the cuisine that the individual is eating, but also identify the specific dish that is being consumed. Our approach hinges on identifying the restaurant the individual is at, and retrieving the menu of said restaurant. Although latitude and longitude can be reliably obtained with GPS sensors in mobile and wearable devices today, there might be times when the association between location data and the exact restaurant the person is visiting is erroneous (e.g. person is inside a shopping mall, or when two or three restaurants are in close proximity to each other). Although this might seem like a limitation of our method, it is usually not of practical concern since restaurants that are physically close are typically significantly different in their offerings. Thus, it is often enough to identify the general physical area the individual is at (as opposed to the exact restaurant) and retrieve the menu of all restaurants and their respective food photos. 

\paragraph{Semi-Automating Food Journaling}

Dietary self-monitoring is effective when individuals are actively engaged and become aware of their eating behaviors. This, in turn, can lead to reflection and modifications in food habits. Our approach to food recognition is designed to facilitate dietary self-monitoring. Engagement is achieved by having individuals take a picture of their food; the tedious and time-consuming task of obtaining details about the food consumed is automated.

\section{Conclusion}

Although numerous solutions have been suggested for addressing the problem of poor adherence to nutrition journaling, a truly practical system for dietary self-monitoring remains an open research question. In this paper, we present a method for automatically recognizing foods eaten in restaurants leveraging location sensor data and online databases.

The contributions of this work are (1) an automatic workflow where online resources are queried with contextual sensor data (e.g., location) to assist in the recognition of food in photographs.; (2) image classification using the SMO-MKL multi-class SVM classification framework with features extracted using color and point-based algorithms; (3) an in-the-wild evaluation of our approach with food images taken in 10 restaurants across 5 different types of food (American, Indian, Italian, Mexican and Thai); and (4) a comparative evaluation focused on the effect of location data in food recognition results.

\textbf{}\\
\textbf{Acknowledgements} Funding and sponsorship was provided by the U.S. Army Research Office (ARO) and Defense Advanced Research Projects Agency (DARPA) under Contract No. W911NF-11-C-0088 and the Intel Science and Technology Center for Pervasive Computing (ISTC-PC). Any opinions, findings and conclusions or recommendations expressed in this material are those of the authors and do not necessarily reflect the views of our sponsors.

{
\bibliographystyle{ieee}
\bibliography{all,food}
}

\end{document}